\documentclass{article}
\usepackage{spconf,amsmath,epsfig}
\usepackage{booktabs}
\usepackage{float}  
\usepackage{graphicx}
\usepackage{subcaption}
\usepackage{xurl}

\usepackage[pagebackref,breaklinks,colorlinks]{hyperref}

\hypersetup{
    colorlinks=true,
    linkcolor=black,
    filecolor=magenta,      
    urlcolor=black,
    }

\newcommand{\paragraphc}[1]{\smallskip\noindent\textbf{#1}}

\title{BlessemFlood21: Advancing Flood Analysis with a High-Resolution Georeferenced Dataset for Humanitarian Aid Support}

\multiauthors
    {Vladyslav Polushko \textsuperscript{1, 2}, Alexander Jenal \textsuperscript{3}, 
    Jens Bongartz \textsuperscript{3}, 
    Immanuel Weber \textsuperscript{3}, 
    Damjan Hatic \textsuperscript{1}, 
    Ronald Rösch \textsuperscript{1}, 
    Thomas März \textsuperscript{2}, 
    Markus Rauhut \textsuperscript{1},
    Andreas Weinmann \textsuperscript{2}}
    {\ninept 1 Image Processing Department, Fraunhofer ITWM, Kaiserslautern, Germany 
    \\[3 pt] 
    \ninept 2 Working Group Algorithms for Computer Vision, Imaging and Data Analysis, Hochschule Darmstadt, Darmstadt, Germany 
    \\[3 pt]
    \ninept 3 Center for Machine Learning and Sensor Technology, Hochschule Koblenz, Remagen, Germany}

\begin{document}

\maketitle

\begin{abstract}
Floods are an increasingly common global threat, causing emergencies and severe damage to infrastructure.
During crises, organisations such as the World Food Programme use remotely sensed imagery, typically obtained through drones,
for rapid situational analysis to plan life-saving actions.
Computer Vision tools are needed to support task force experts on-site in the evaluation
of the imagery to improve their efficiency and to allocate resources strategically.
We introduce the BlessemFlood21 dataset
to stimulate research on efficient flood detection tools.
The imagery was acquired during the 2021 Erftstadt-Blessem flooding event and
consists of high-resolution and georeferenced RGB-NIR images. In the resulting RGB dataset,
the images are supplemented with detailed water masks, 
obtained via a semi-supervised human-in-the-loop technique, where in particular the NIR information
is leveraged to classify pixels as either water or non-water. 
We evaluate our dataset by training and testing 
established Deep Learning models for semantic segmentation.
With BlessemFlood21 we provide labeled high-resolution RGB data and a baseline for further development of algorithmic solutions tailored to flood detection in RGB imagery.

\end{abstract}
\section{Introduction}
\label{sec:intro}
Over the past decade, there has been a rise in both the frequency and severity of flooding events, 
leading to significant human and economic losses. 
From 2000 to 2020, there were 163 floods annually around the world, making it the most frequent natural disaster~\cite{disastercred21, swiss21, ourworld2021online, IPPC21}. 
Rapid humanitarian response, medical assistance, and rescue, mitigate the impact of floods.
Nowadays, drones are used to capture images of the flooded areas
which are evaluated to create inundation maps~\cite{gebrehiwot19dlmappingdroneflood} 
as well as vulnerability maps for future risk analysis~\cite{UnicefDronesFlood22, karamuz20dronesfloodhazardassessment}. 
On-site emergency responders face time constraints.
Therefore, algorithms that perform Computer Vision tasks for natural disaster analysis \cite{gebrehiwot20inundationmappingndwidl, RealTimeFloodDetectionDL21, HumanitarianDronesFloodDeepLearning20, karamuz20dronesfloodhazardassessment} 
efficiently are crucial to support emergency responders.  
Traditionally, vegetation indices have been used to analyse flood-prone regions \cite{gebrehiwot20inundationmappingndwidl} 
and to detect water in natural disaster scenarios.
More recently, Deep Learning (DL) methods for semantic segmentation have been employed to detect water in flood imagery
and are able to accurately identify and delineate flooded areas \cite{hashemi21floodmappinguav}.
This ability enables precise identification of damaged infrastructure to support task force experts.
To obtain accurate water masks, 
it is essential to train semantic segmentation models on high-quality annotated imagery of real-life flood scenarios~\cite{Spacenet8Hansch22,floodnet21, furukawa21dronemsmoreexpensivedronergb, karamuz20dronesfloodhazardassessment}.
Drones equipped with RGB cameras are inexpensive and allow for the reliable capture of high-resolution images 
within a range of ground sampling distance depending on their flight altitude and camera resolution~\cite{halbgewachs23usinguavmappingstructuraldamage, hashemi21floodmappinguav}. 
Having high-resolution imagery at your disposal is crucial~\cite{zheng18geofloodterrainhighresolutionneeded} because it provides
valuable contextual information for improved situational analysis.
To date, there is only one dataset, FloodNet~\cite{floodnet21}, that provides high spatial resolution RGB-imagery captured by drones. 
FloodNet depicts coastal flooding scenes at 1.5~cm spatial resolution, but lacks full nadir acquisition and is confined to a single area.
Other flooding datasets are available, but have lower spatial resolution:  
SpaceNet~8~\cite{Spacenet8Hansch22} relies on satellite imagery to cover a vast area. The resulting dataset consists of RGB data upsampled to a resolution of 30~cm, but does not provide segmentation masks.
xBD~\cite{xbdgupta19} provides RGB~imagery of post-flood landscapes at lower spatial resolution of 0.8~m. 
It covers a substantial area, provides labeled imagery exclusively for building damage detection, and contains limited flooded imagery. 
Likewise, SEN12-FLOOD~\cite{rambour20Sen12flood} is a multimodal dataset for flood scene classification, and hence does not provide segmentation masks. 
This discussion underscores the critical need for additional high-resolution RGB datasets. Specifically, there is a significant gap in datasets that focus on non-coastal river scenes. Such datasets are essential for semantic segmentation purposes, particularly in detecting water bodies and mapping flooded regions accurately.

\paragraphc{Contribution.}
This paper introduces the BlessemFlood21 dataset for semantic water segmentation, 
providing RGB-imagery at 15~cm spatial resolution of non-coastal flooded river scenes. 
The imagery was acquired via a gyrocopter in a nadir setup in the aftermath of the Ahrtal flood in July 2021.  
In particular, our contributions are as follows:
\begin {itemize}
\item[(i)] We provide imagery of Blessem shortly after the flood event, which depicts, in particular, multiple rivers in a non-coastal scenario. We add to the publicly available datasets a novel variety of flood sceneries.

\item[(ii)] 
We provide annotated data for semantic RGB-based segmentation of water. To this end, we employ a semi-automatic human-in-the-loop strategy and leverage data from an additional NIR channel provided by our imaging setup. We also demonstrate the benefits of NIR band information w.r.t. creating water masks.

\item[(iii)]
We evaluate the dataset with three established Deep Learning models to provide a baseline for further flood analysis. 
\end{itemize}
With these contributions, we aim to lay the groundwork  to tackle urgent issues in flood analysis, thereby
providing assistance to both researchers and humanitarian aid organisations.

\section{Dataset}

In response to the devastating flooding in Erftstadt, Germany, in July 2021, a dedicated aerial survey mission was carried 
out to acquire imagery of the affected areas. Amidst multiple flights for rescue and mapping, this mission aimed for gathering 
detailed data of the Blessem district in the city of Erftstadt with a specific emphasis on the edge of a gravel pit.

\paragraphc{Acquisition.}
The gyrocopter was chosen for the aerial survey due to its extended flight capability, allowing for several hours of flight time and 
supporting a payload of up to 100 kg. Gyrocopters have the ability to operate at an altitude as low as 150~m at a reduced speed of 30 km/h, 
allowing for a smooth acquisition process. 
For BlessemFlood21, a single 80-minute flight is conducted, with the gyrocopter maintaining a ground speed of 100~km/h and operating at an acquisition altitude of 522~m.

\paragraphc{Imagery.}
The imagery was acquired using the PanX.3 multi-camera system, developed by Jenal et al.~\cite{jenal22adaptive}. 
The system is attuned to the visible and near-infrared (VNIR) spectral range (400-900 nm) and incorporates four Sony IMX304 sensors, 
allowing for image capturing at a resolution of $4112 \times 3008$ pixels. 
By pairing a sensor with a 12~mm focal length lens, a high spatial resolution of 15~cm per pixel was achieved due to the low altitude. 
In addition, WGS84 (EPSG:4326) coordinates and their corresponding projections were recorded by a GNSS sensor, facilitating accurate georeferencing. 
The images were processed to produce a georeferenced multispectral orthomosaic. The mosaic obtained has a resolution of $35,701 \times 34,508$ pixels and provides standard RGB channels and an extra 
NIR channel, saved in geotiff format. 
We note that the NIR channel is used for annotation, while the resulting annotated dataset is RGB-only.
The dataset is available at:
\url{https://fordatis.fraunhofer.de/handle/fordatis/379}


\section{Annotation Framework}
In our annotation process, the aim is to create segmentation masks that label each pixel as either water or non-water. However, manual labelling presents challenges due to the large number of pixels. To overcome this, we use classical Machine Learning and employ a Random Forest classifier~\cite{breiman01randomforest} to improve accuracy and speed up the process. Our solution includes a human-in-the-loop framework that comprises preprocessing, human annotators, Machine Learning-driven segmentation, and visual assessment. 
During annotation, we set the following criteria: water bodies, flooded regions, and lakes are labeled as water, including their residuals. Conversely, buildings, roads, vehicles, and mud are labeled as non-water.
The RGB-NIR orthomosaic is divided into 4623 tiles, each measuring 512 x 512 pixels, to overcome hardware limitations. To address low contrast of the tiles, we normalize pixel values within the 2nd and 98th percentiles of the mosaic. 
Next, we transform the raw tiles into false-colour RGB tiles. For this, we explore combinations of three channels, ensuring that each combination includes the NIR channel, as it is particularly helpful for highlighting water features owing to its characteristic light absorption within the NIR spectrum \cite{ahn20estimatingwaterabsorbsnir}.  
Empirical evaluation reveals the near infrared (NIR), blue (B) and red (R) channels as the most effective combination.
\begin{figure}[t]
  \centering
  \includegraphics[width=\columnwidth]{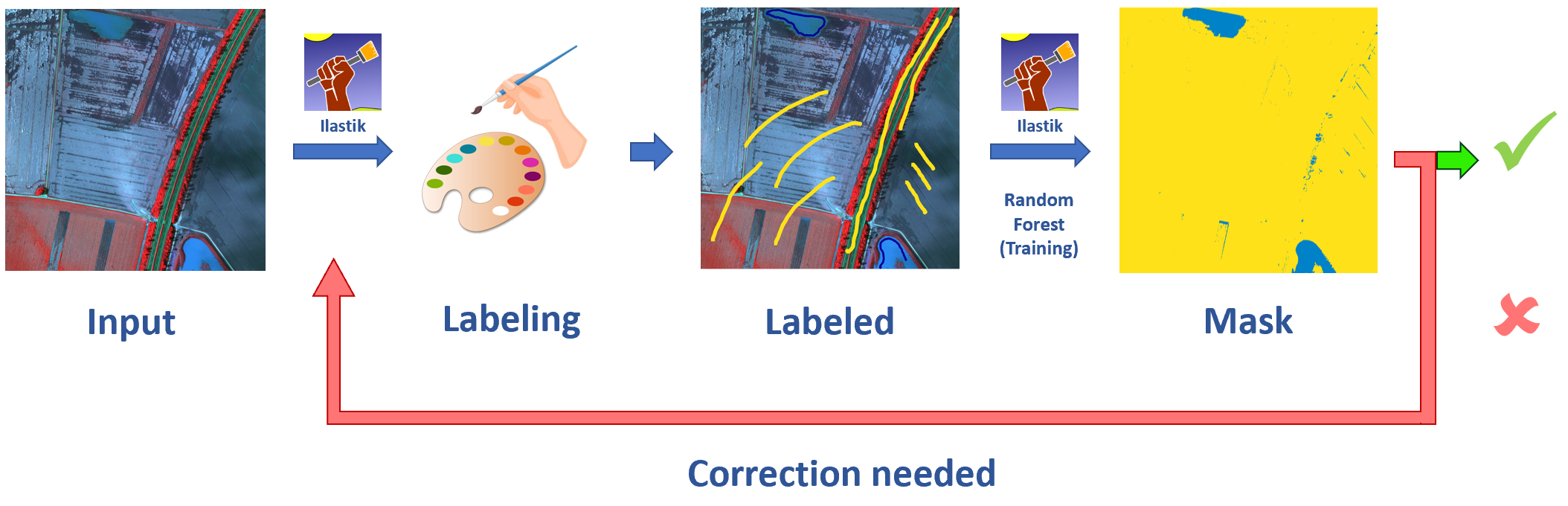}
  \caption{Human-in-the-loop framework used for generation of ground truth masks.}
  \label{fig:human_in_the_loop}
\end{figure}
Next, our workflow integrates Ilastik \cite{berg19ilastik} for annotation and segmentation. This toolkit incorporates interactive supervised Machine Learning with sparse labels, which significantly improves both speed and accuracy. 
During water annotation, a subset of false-colour RGB tiles is labeled by the human-in-the-loop with sparse brush strokes, highlighting the boundaries along flooded regions and rivers as they represent the highest segmentation uncertainty. 
Using Ilastik's pixel classification pipeline, we apply a Random Forest classifier~\cite{breiman01randomforest} with 100 trees for segmentation.
\begin{figure}[!htb]
  \centering
  \begin{subfigure}[b]{\linewidth}
    \centering
    \includegraphics[width=0.42\linewidth]{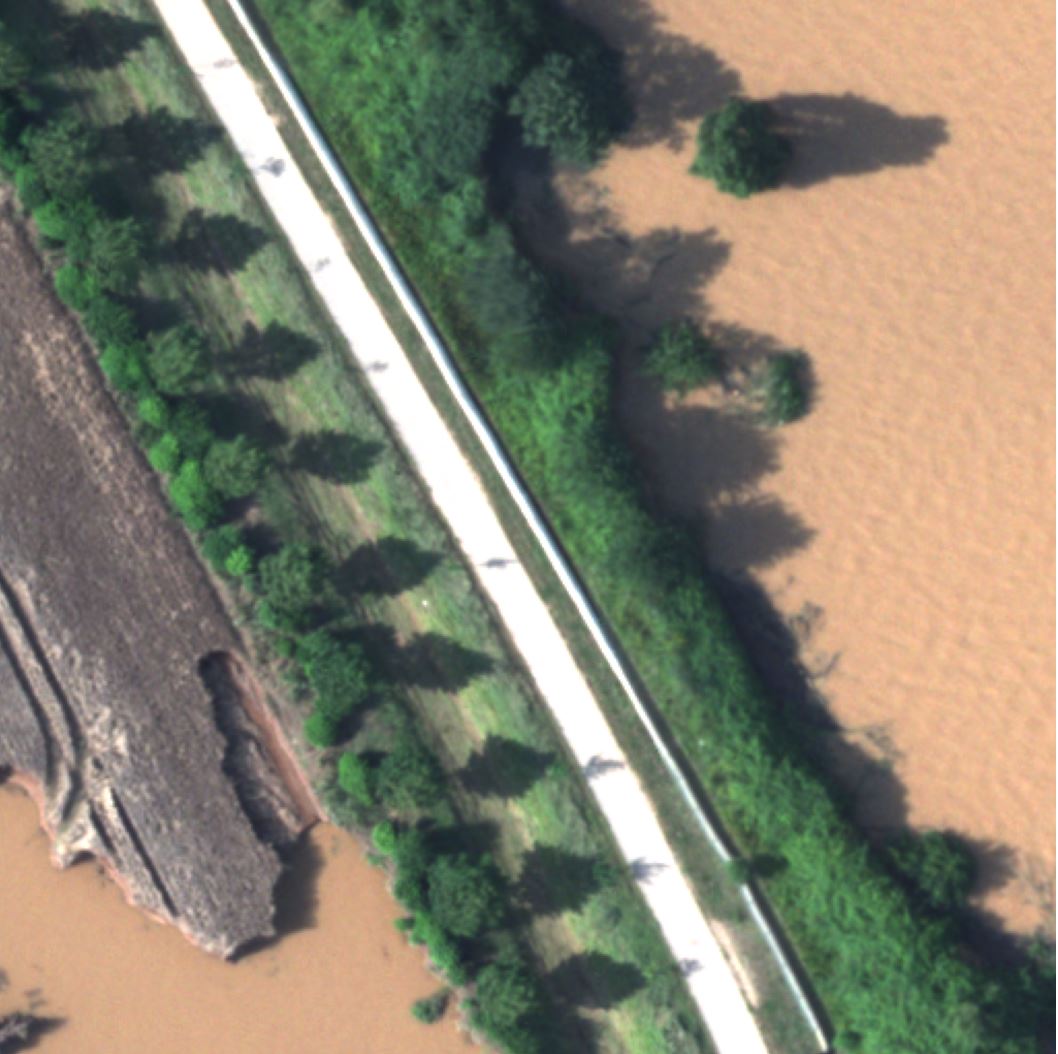}
    \hspace{3mm}
    \includegraphics[width=0.42\linewidth]{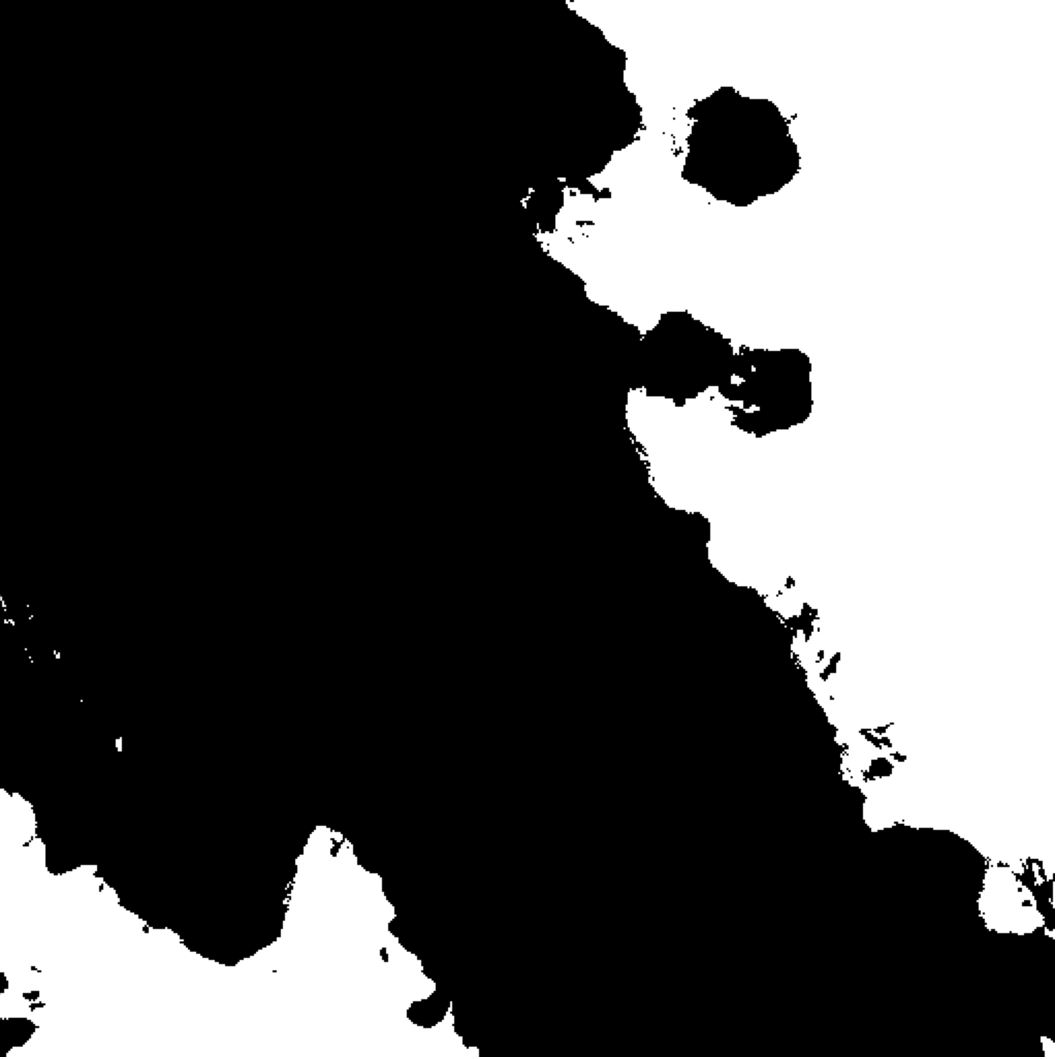}
 \caption{Rural damage on crop field.}
  \end{subfigure} 
 
  \begin{subfigure}[b]{\linewidth}
    \centering
    \includegraphics[width=0.42\linewidth]{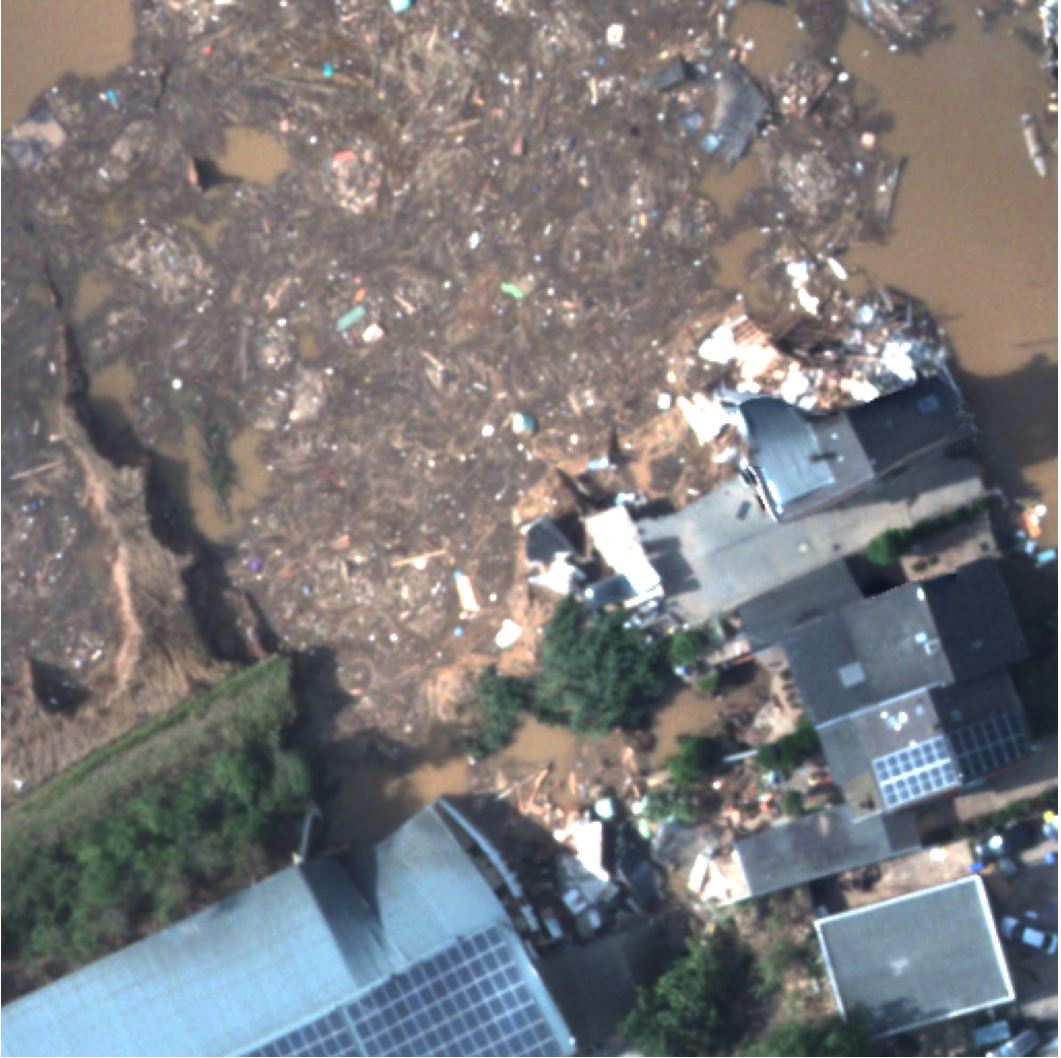}
    \hspace{3mm}
    \includegraphics[width=0.42\linewidth]{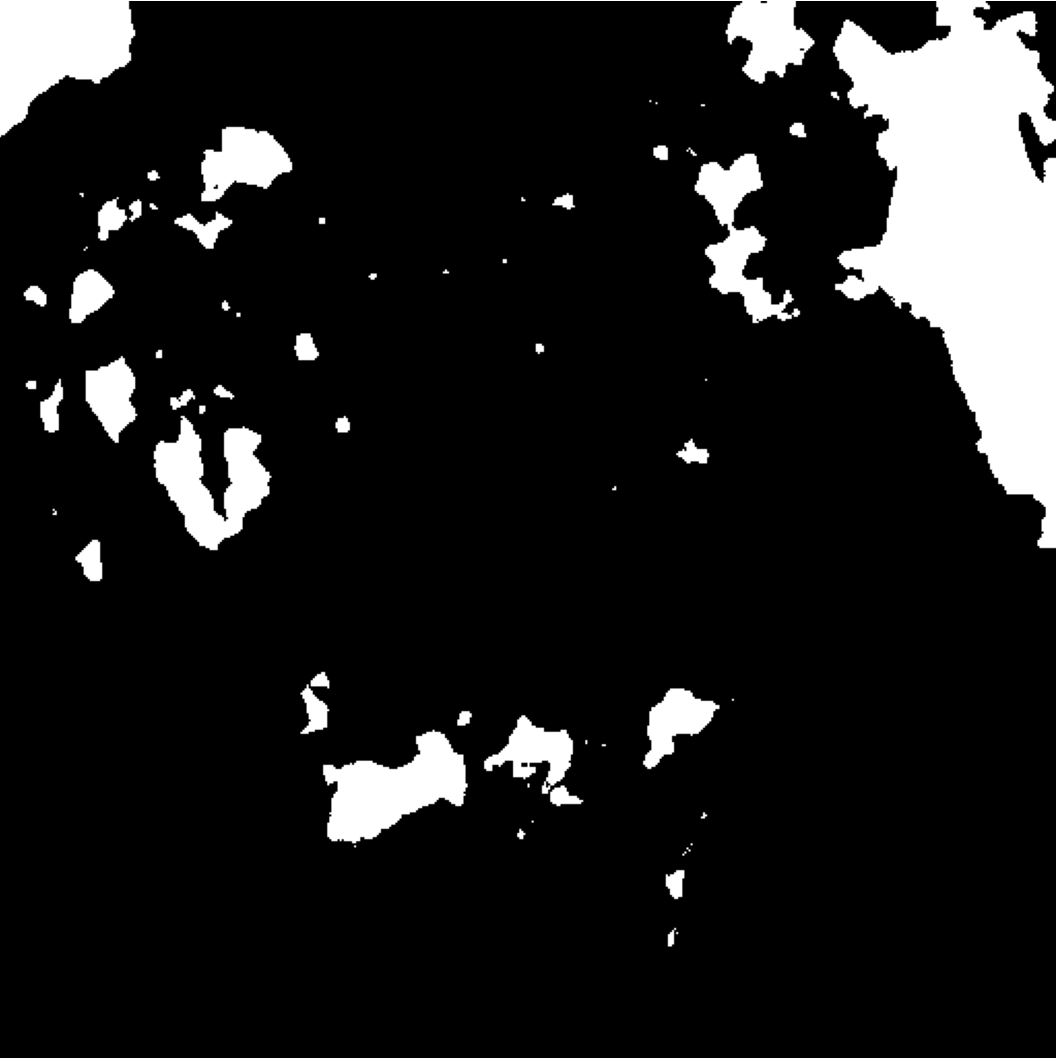}
    \caption{Urban damage with flooded infrastructure and debris.}
  \end{subfigure}

  \begin{subfigure}[b]{\linewidth}
    \centering
    \includegraphics[width=0.42\linewidth]{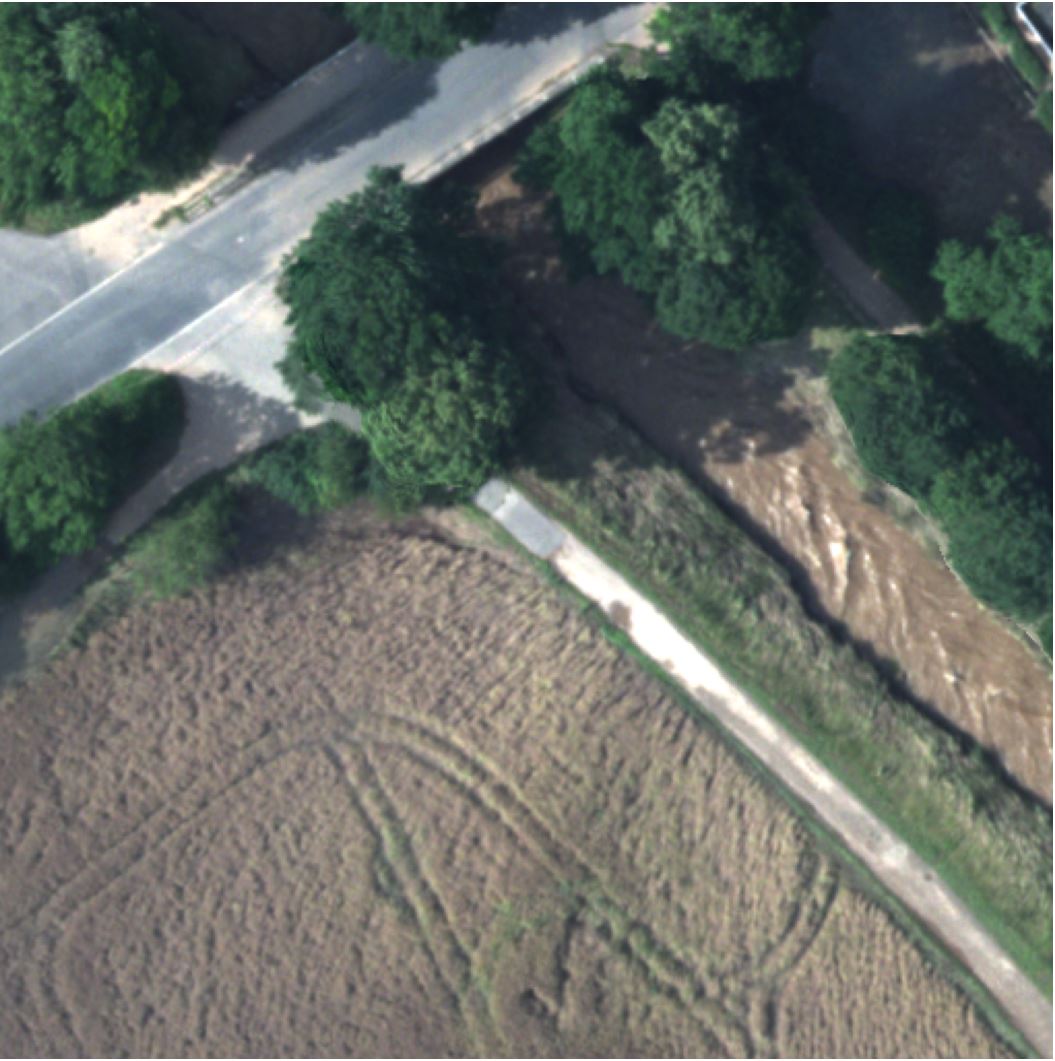}
      \hspace{3mm}
     \includegraphics[width=0.42\linewidth]{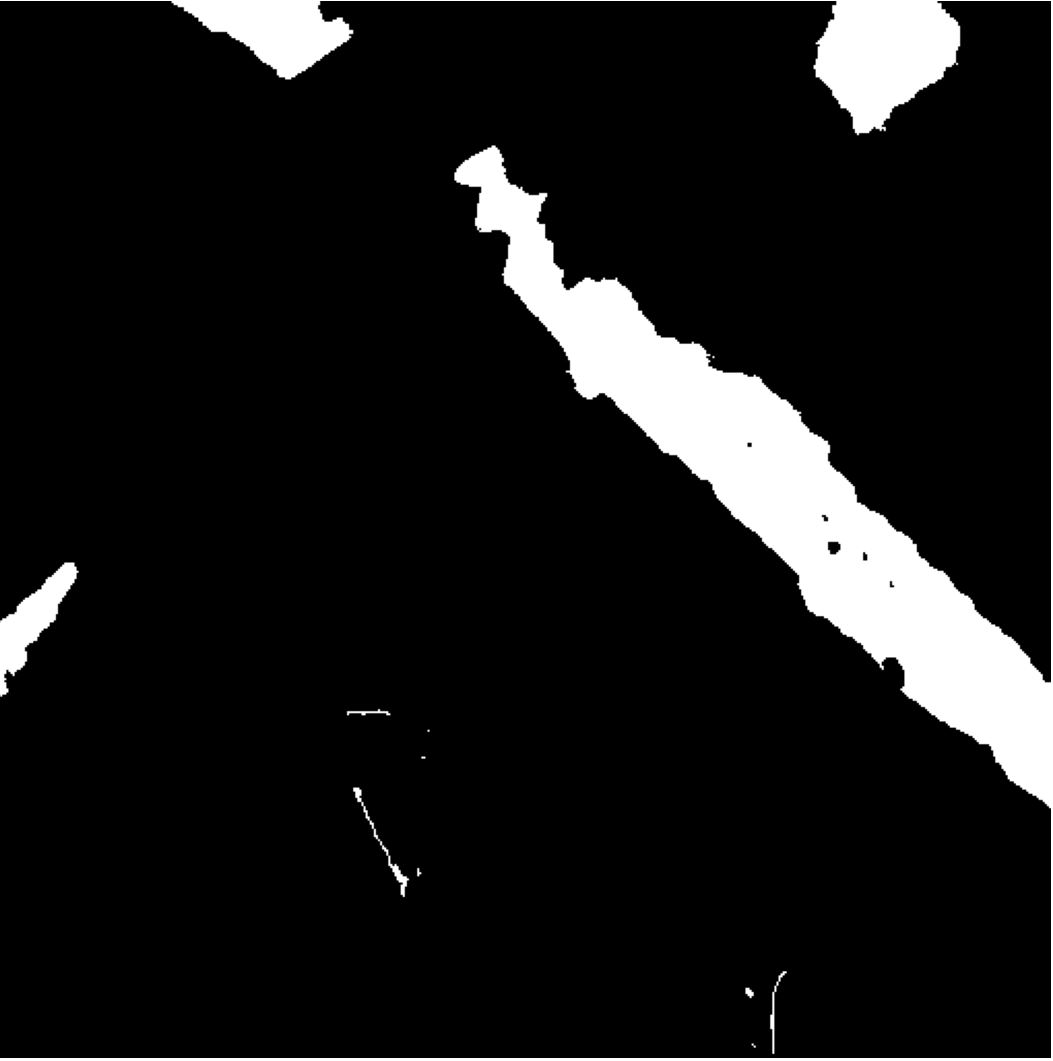}
  \caption{River with turbid water and floodwater residuals.}
  \end{subfigure}

  \caption{Three representative sample pairs of preprocessed imagery (left) and generated ground truth water masks (right). White pixels in the masks indicate water pixels.}
  \label{fig:Generate Ground Truth Samples}
\end{figure}

Evaluating segmentation results using metrics is not feasible due to the lack of accurate ground truth. 
To address this issue, we rely on a visual comparison between false colour and true colour tiles. This method supports decision-making in difficult scenarios where differentiating between mud and muddy water is challenging. 
Initial segmentations provide mostly accurate water shapes, but further improvements are necessary. To improve mask quality, a human-in-the-loop refinement process is used. Any mask rejected by the human annotator triggers additional labelling and iterative model retraining. This process is repeated until the masks meet the desired accuracy (see Figure \ref{fig:human_in_the_loop}).
After generating the ground truth masks, georeferences are transferred from their raw tile counterparts.
Finally, we provide the georeferenced mosaic along with its corresponding binary water mask. In response to real-world humanitarian aid scenarios, we exclude non-RGB information, aligning with practical limitations.

\section{Baseline}
\label{sec:baseline}

We establish a baseline for model performance and lay the foundation for future research. We highlight the suitability of our dataset for DL algorithms, emphasising their critical role in ensuring reliable water detection results.
For our baseline approach, we leverage three state-of-the-art semantic segmentation models: DeeplabV3+~\cite{chen18deeplabv3+}, UNet++~\cite{zhou18unet++}, and SegFromer~\cite{xie21segformer}.  
DeeplabV3+ excels in capturing fine details, UNet++ is known for its excellent boundary delineation, and SegFormer offers a novel transformer-based approach, providing a diverse baseline for further research. 
Through the application of different model architectures, we provide an assessment of the applicability of our dataset in different architectural contexts.
We preserve fine details by using 4623 tiles, each 512 by 512 pixels. We divide the BlessemFlood21 dataset into 80\% training, 10\% validation, and 10\% test sets. Specifically allocating 10\% of the water-containing images to both the validation and test sets to ensure diversity across these subsets.
For improved model generalisation, we use basic augmentations from the Albumentations library~\cite{buslaev20albumentations}, such as horizontal and vertical flipping, rotation, translation, scaling and random cropping.
We train on an NVIDIA A100 with 40 GB running 100 epochs, ensuring efficient training and evaluation using a PyTorch implementation.

\section{Results and Discussion}

\begin{figure*}[t]
  \centering
  \begin{subfigure}{0.18\linewidth}
    \includegraphics[width=\linewidth]{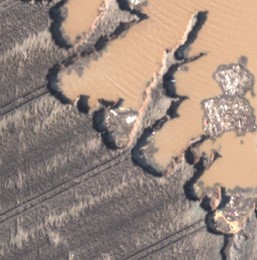}
    \caption{Original}
    \label{fig:subfig1}
  \end{subfigure}
  \hfill
  \begin{subfigure}{0.18\linewidth}
    \includegraphics[width=\linewidth]{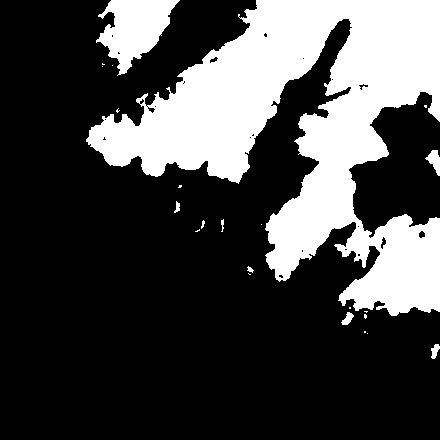}
    \caption{Ground Truth}
    \label{fig:subfig2}
  \end{subfigure}
  \hfill
  \begin{subfigure}{0.18\linewidth}
    \includegraphics[width=\linewidth]{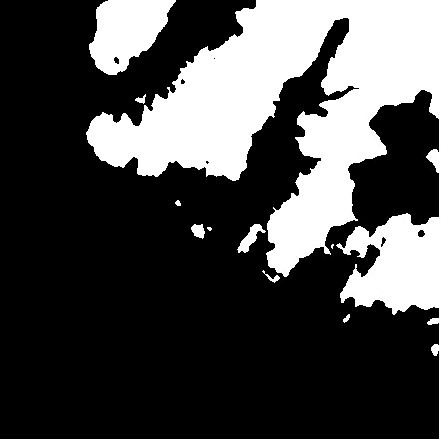}
    \caption{UNet++}
    \label{fig:subfig3}
  \end{subfigure}
  \hfill
  \begin{subfigure}{0.18\linewidth}
    \includegraphics[width=\linewidth]{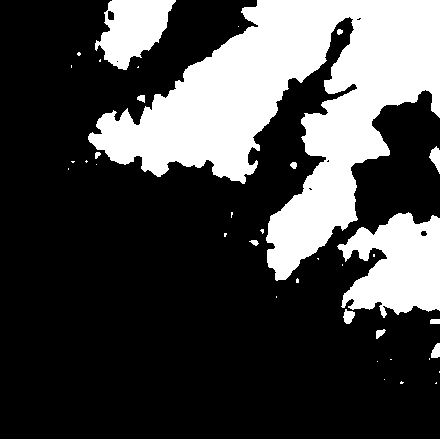}
    \caption{DeepLabV3+}
    \label{fig:subfig4}
  \end{subfigure}
  \hfill
  \begin{subfigure}{0.18\linewidth}
    \includegraphics[width=\linewidth]{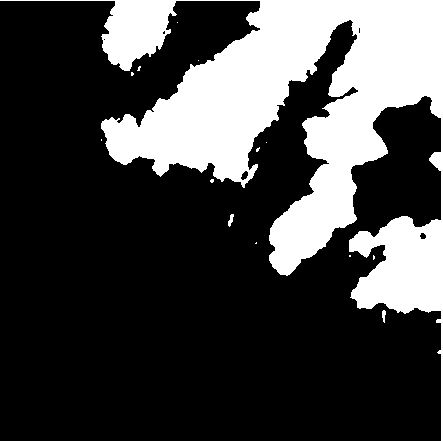}
    \caption{SegFormer}
    \label{fig:subfig5}
  \end{subfigure}

  \caption{Qualitative comparison of ground truth and predictions of water masks obtained after fine-tuning with 100 epochs.}
  \label{fig:Intermodel Comparison}
\end{figure*}

\paragraphc{Dataset Quality Assessment.}
We have generated a dataset for water segmentation.
To label the data, we have developed a human-in-the-loop strategy; we rely on additionally employing a NIR channel provided by our imaging setup. 
The accuracy of the generated data is assessed by visual inspection. Due to the fact that the lack of ground truth prevents the use of conventional metrics.
We display representative samples in
Figure \ref{fig:Generate Ground Truth Samples}. 
Figure \ref{fig:Generate Ground Truth Samples}a shows rural damage caused by flooding.
We see a crop field and a newly formed edge as the result of a landslide.
Figure \ref{fig:Generate Ground Truth Samples}b is an example of an urban environment near a gravel pit. 
As floodwater residuals, we observe partially flooded buildings and scattered furniture. 
Finally, Figure \ref{fig:Generate Ground Truth Samples}c shows a river carrying floodwater.
The examples both illustrate the complexity of floodwater detection
and the high quality of the generated masks obtained by
employing our human-in-the-loop approach.
We emphasise that the presence of shadows and the turbidity of floodwater make the accurate 
distinction of floodwater areas from their surroundings challenging.
In particular, the floodwater near the bridge in Figure~\ref{fig:Generate Ground Truth Samples}c is well detected despite the shadows of the trees,
which demonstrates the ability of the framework to handle inhomogeneous lighting conditions.
Moreover, the framework delineates complex boundaries of flooded areas, both of the landscape (Figure~\ref{fig:Generate Ground Truth Samples}a) and of urban flood residues (Figure~\ref{fig:Generate Ground Truth Samples}b).   
To precisely capture complex boundaries is fundamental for the accurate characterisation of flood-induced damages. 
It is important to note that the resulting dataset exhibits a bias towards the non-water class, with about 85\% of pixels belonging to this category. 
This bias can be primarily attributed to the prevalence of river scenes. These scenes are important since
the accurate depiction of rivers experiencing elevated water levels is valuable for flood analysis. 
Summing up, the proposed dataset provides data with water masks of high accuracy showing flooded landscapes with complex boundaries. 
The dataset provides detailed insights into spatially diverse flood scenarios, with a particular focus on river inundation near urban cities. 
Its precision can pave the way for a more profound understanding of flood dynamics.

\paragraphc{Model Performance.}
We evaluate the performance of the three baseline models DeepLabV3+, UNet++, and SegFormer 
(cf. Section~\ref{sec:baseline}).
Figure \ref{fig:Intermodel Comparison} shows the corresponding segmentation results. 
All networks successfully detect coarse scale features, smooth boundary delineation, as well as turbid and muddy water.  
However, the models fail to capture fine-scale details represented by smaller soil residuals. 
Small-scale objects enclosed by muddy water are often either undetected or 
incorrectly classified as water (see the centre of the predicted masks in Figure~\ref{fig:Intermodel Comparison}).
\begin{table}[t]
\centering
\begin{tabular}{p{1.8cm}p{2.cm}p{1.cm}p{0.8cm}p{0.8cm}}
\toprule
Model & Encoder & Params & IoU (\%) & Dice (\%) \\
\midrule
DeepLabV3+  & Resnet-50 & 26.7 M &  90.6 & 95.1 \\ 
UNet++      & Resnet-50 & 49.0 M &  84.2 & 91.4 \\ 
SegFormer-b5& Hierarchical ViT & 84.6 M &  75.4 & 78.4 \\ 
\bottomrule
\end{tabular}
\caption{Performance comparison of water detection models on the BlessemFlood21 test set.}
\label{tab:model_comparison_blessem}
\end{table}
For quantitative evaluation, we compare the models on the BlessemFlood21 test set using
Intersection over Union (IoU) and the Dice Score (Dice). 
Table~\ref{tab:model_comparison_blessem} displays the results.
All models yield IoU values higher than $75\%$ and Dice scores higher than $78\%$.
In summary, the baseline methods both quantitatively and qualitatively demonstrate good performance in water segmentation on the Blessem21Flood dataset. 
We observe that there is room for improvement w.r.t. the segmentation of small-scale details.

\paragraphc{Comparison with FloodNet.} 
Finally, we compare the BlessemFlood21 dataset, labeled with our framework, with the FloodNet dataset.
We chose FloodNet because its acquisition setup is similar to BlessemFlood21. FloodNet captures flooded regions using high-resolution imagery, similar to BlessemFlood21. However, FloodNet focuses on hurricane-induced flooding in the USA, with greater water coverage and diverse vegetation such as palm trees. In contrast, BlessemFlood21 concentrates on river-induced floods in Germany following heavy rains.
To underpin the difference between these datasets, we made the following experiment:
we fine-tuned the models on FloodNet and evaluated their performance on BlessemFlood21.
We observe that the models exhibit poor performance when applied to BlessemFlood21: the highest overall IoU was only $10.5\%$ and the highest Dice Score was $19.0\%.$ 
Based on that, we underline the unique features of our dataset and its potential impact on future research. 


\section{Conclusion}
We introduced the BlessemFlood21 dataset for water segmentation. The dataset consists of RGB images 
with a spatial resolution of 15 cm. It depicts 
post-flood scenes involving mainly (non-coastal) river scenes for semantic segmentation. 
We have provided image data of Blessem after the food event.
Further, we have annotated the raw image material using an iterative human-in-the-loop annotation framework together with an additional NIR channel provided by our imaging setup. 
As a result, we obtained a labeled dataset for semantic RGB-based segmentation of water. 
Finally, we have evaluated three state-of-the-art Deep Learning segmentation models on our dataset.
We observed the top performance by DeepLabV3+ yielding 
an IoU of $90.6\%$ and a Dice Score of $95.1\%.$ 
Compared to FloodNet, which is in a sense the closest related dataset, we observed the added value of the proposed dataset.

\bibliographystyle{IEEEbib}
\bibliography{strings}

\end{document}